\pdfoutput=1
\documentclass[11pt]{article}

\usepackage[]{acl}

\usepackage{times}
\usepackage{latexsym}
\usepackage{enumitem}

\newcommand{\alignsum}{\textsc{AlignSum}}

\usepackage[T1]{fontenc}
\usepackage[utf8]{inputenc}

\usepackage{microtype}
\usepackage{graphicx} %插入图片的宏包
\usepackage{float} %设置图片浮动位置的宏包
\usepackage{subfigure} %插入多图时用子图显示的宏包

\usepackage{amssymb}

\usepackage{multirow}
\usepackage{multicol}
\usepackage{diagbox}
\usepackage{graphicx}
\usepackage{colortbl}
\usepackage{algorithm}
\usepackage{algorithmic}
\usepackage{amsmath}
\usepackage{hyperref}
\usepackage{url}
\usepackage{booktabs}
\usepackage{xcolor}
\title{\alignsum: Data Pyramid Hierarchical Fine-tuning for \\ Aligning with Human Summarization Preference}

\author{Yang Han$^{1,2,3}$\thanks{\ \ Yang Han and Yiming Wang contribute equally.} ~~Yiming Wang$^{2*}$ ~~Rui Wang$^{2\dagger}$ ~~Lu Chen$^{1,2,3}$ ~~Kai Yu$^{1,2,3}$\thanks{\ \ Rui Wang and Kai Yu are the corresponding authors.} \\
$^{1}$X-LANCE Lab, Department of Computer Science and Engineering, SJTU \\
$^{2}$ MoE Key Lab of Artificial Intelligence, SJTU AI Institute \\Shanghai Jiao Tong University, Shanghai, China \\
$^{3}$Suzhou Laboratory, Suzhou, China \\
{\tt \{csyanghan,wangrui12,kai.yu\}@sjtu.edu.cn}}

\begin{document}
\maketitle

\begin{abstract}
% 一个很大的扩展就是：能不能根据某种算法选择某个样本应该采用Abstractive data还是extractive data
% 
% 
Text summarization tasks commonly employ Pre-trained Language Models (PLMs) to fit diverse standard datasets. While these PLMs excel in automatic evaluations, they frequently underperform in human evaluations, indicating a deviation between their generated summaries and human summarization preferences.
This discrepancy is likely due to the low quality of fine-tuning datasets and the limited availability of high-quality human-annotated data that reflect true human preference.
To address this challenge, we introduce a novel human summarization preference alignment framework {\bf \alignsum}. This framework consists of three parts: Firstly, we construct a Data Pymarid with extractive, abstractive, and human-annotated summary data. Secondly, we conduct the Gaussian Resampling to remove summaries with extreme lengths. Finally, we implement the two-stage hierarchical fine-tuning with Data Pymarid after Gaussian Resampling.
We apply \alignsum\ to PLMs on the human-annotated \textit{CNN/DailyMail} and \textit{BBC XSum} datasets. Experiments show that with \alignsum, PLMs like B{\footnotesize ART}-Large surpass 175B GPT-3 in both automatic and human evaluations. This demonstrates that \alignsum\ significantly enhances the alignment of language models with human summarization preferences.\footnote{The code is released at: \url{https://github.com/csyanghan/AlignSum}}

% This method involves three tiers of data, characterized by decreasing quantity and increasing quality. The initial two tiers form the base, providing the bulk of the training data and enhancing the PLMs' capacity to generate domain-general summaries. The top tier, though less abundant, consists of high-quality human-annotated data, without specific instructions. 
% Furthermore, we propose a summarization framework named AlignSum that employs Gaussian resampling and a two-stage training process tailored for PLMs trained on the DP. We assessed AlignSum using the human-annotated \textit{CNN/DailyMail} and \textit{BBC XSum} datasets. Experiments indicate that AlignSum significantly enhances the summarization capabilities of PLMs, such as BART-Large, surpassing the performance of the 175B GPT-3 model in both automatic and human evaluations. 

% The code for our study is accessible at \url{https://github.com/csyanghan/AlignSum}.

\end{abstract}

\section{Introduction}

Text summarization is a pivotal component of natural language processing, striving to produce coherent and concise summaries of textual documents \citep{mani1999advances, nenkova2012survey, allahyari2017text}.
It can be categorized into two main styles: 
{\it Extractive summarization} \citep{nallapati2017summarunner, zhou2020joint, zhong2020extractive} involves selecting significant portions of the text directly from the source; In contrast, {\it abstractive summarization} \citep{see2017get, lewis2019bart} involves generating new text that conveys the original content's essential meaning.
Studies in this field often train Pre-trained Language Models (PLMs) \citep{vaswani2017attention, radford2018improving, lewis2019bart, raffel2020exploring} on standard datasets such as \textit{CNN/DailyMail} \citep{nallapati2016abstractive} and \textit{BBC XSum} \citep{narayan2018don} to fit summary features.
They usually report the performance with reference-based {\it automatic scores} such as R{\footnotesize OUGE} \citep{lin2004rouge}, which directly compare generated summaries with gold summaries, and fine-grained {\it human ratings}, which actually reflect underlying human preferences.

\begin{figure}[t]
    \centering
    \includegraphics[width=0.48\textwidth]{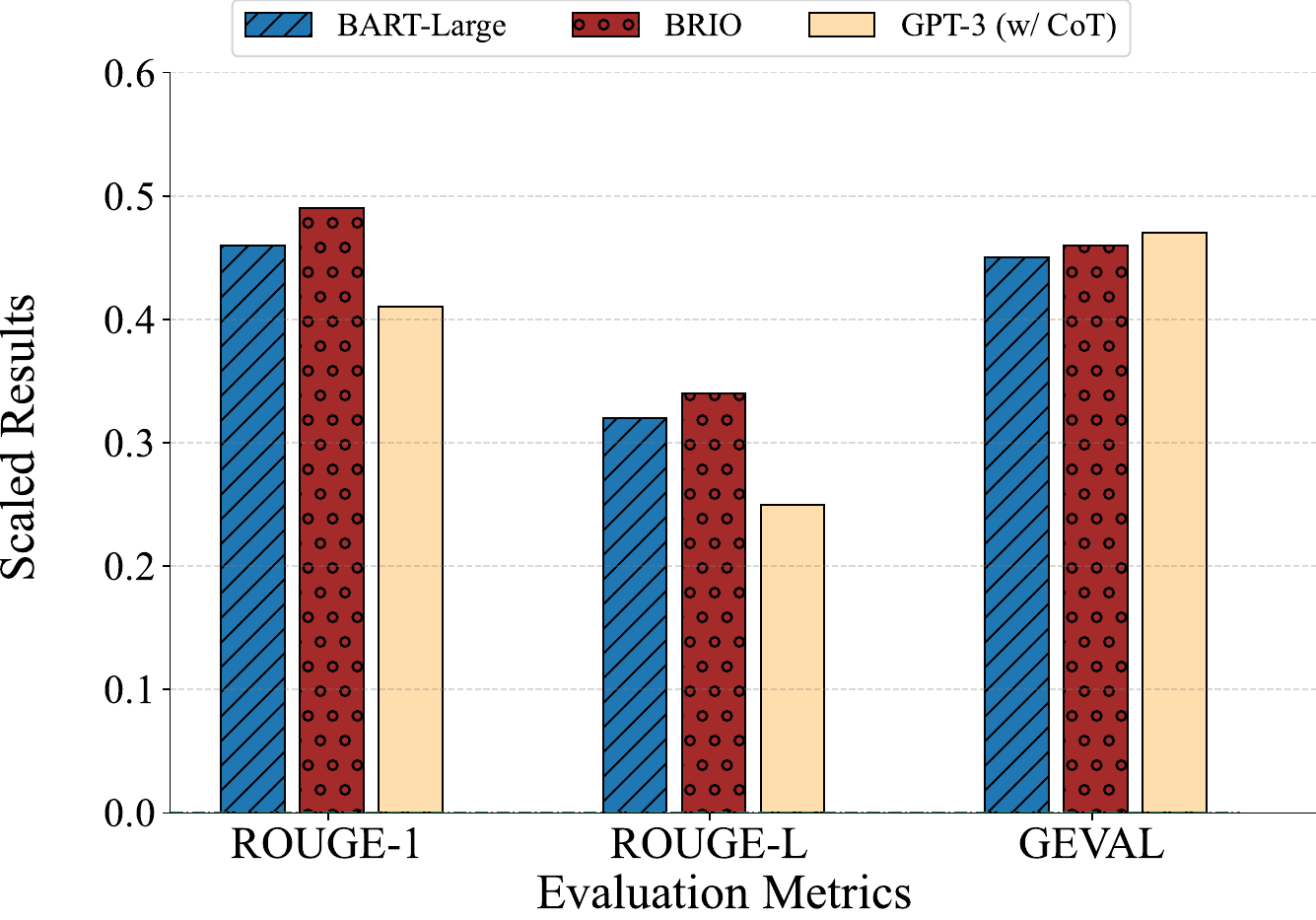}
    \caption{Results (scaled to 0-1) of automatic score R{\footnotesize OUGE} \citep{lin2004rouge} and human rating GEval\footnotemark \cite{liu2023gpteval} on the standard dataset \textit{CNN/DailyMail}. It is obvious that PLMs perform better than LLMs on automatic scores but worse on human ratings.}
    \label{fig:auto_geval}
\end{figure}
\footnotetext{Four aspects: Coherence (score range: 1-5), Consistency (score range: 1-5), Fluency (score range: 1-3), Relevance (score range: 1-5). We use GPT-4 rating that closely aligns with human judgments \citep{liu2023gpteval, wang2023chatgpt}.}

\begin{figure*}[t]
    \centering
    \includegraphics[width=1\textwidth]{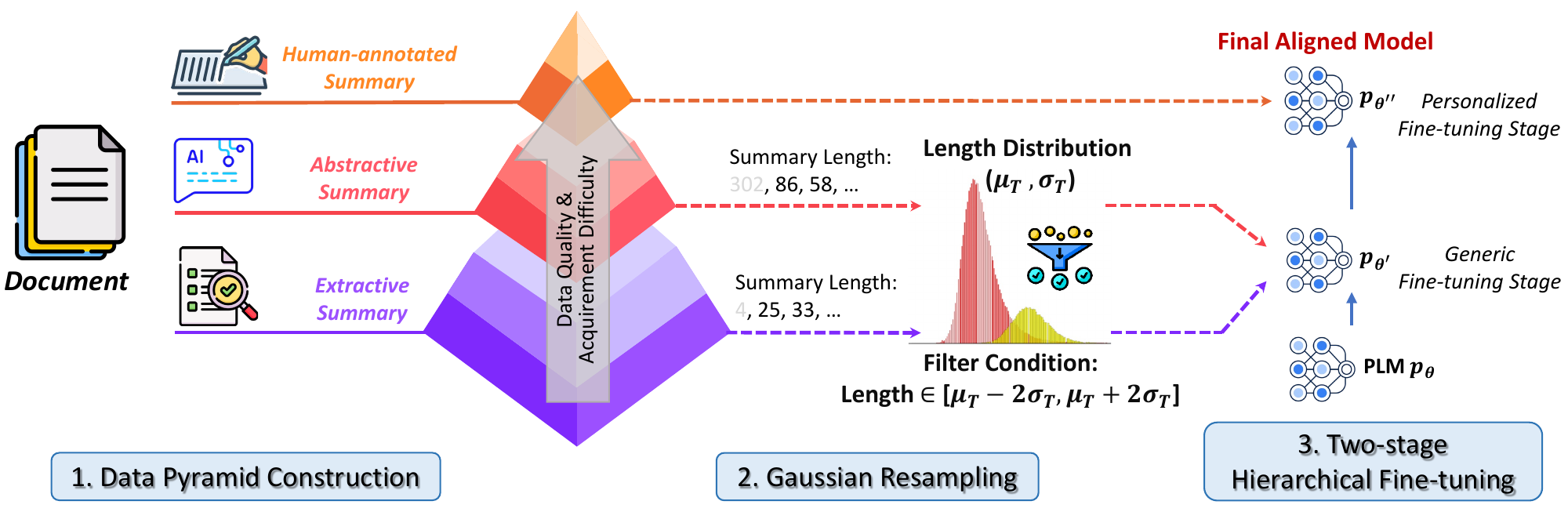}
    \caption{The overall pipeline of our summarization preference alignment framework \alignsum.}
    \label{fig:dp_overall}
\end{figure*}

% However, recent investigations have identified significant issues with the annotation quality of these datasets, such as poor incoherence and factual hallucination \citep{maynez2020faithfulness, kang2020improved, wang2023element}.
% Training PLMs on such datasets would result in generating summaries with high automatic scoring but poor human rating. 

However, recent investigations \citep{goyal2022news,wang2023element} have revealed inconsistencies between automatic scores and human ratings for both PLMs and large language models (LLMs).
As shown in Figure \ref{fig:auto_geval}, when compared to LLMs like GPT-3 (with Chain-of-Thought, CoT) \citep{wang2023element}, PLMs like B{\footnotesize ART}-Large \citep{lewis2019bart} and BRIO \citep{liu2022brio} fine-tuned on the \textit{CNN/DailyMail} demonstrate impressive performances on automatic scores exceeding LLMs, but poor performances on human ratings.
This contradiction stems from that PLMs are fitting low-quality summary data \citep{wang2023element}, indicating that they need more high-quality data for aligning with human preferences to perform better in human ratings.

On the other hand, annotating a large number of high-quality summary datasets is impractical:
(1) Regarding the time cost, the average reading rate for a native English speaker is approximately 220 words per minute \citep{gleni2019assessing, brysbaert2019many}. Moreover, summarization involves a structured cognitive process: reading, comprehending, and summarizing, annotators often spend twenty minutes or more to write a single summary \citep{wang2023element}, rendering the annotation process time-consuming;
(2) Regarding the labor cost, ensuring the accuracy and consistency of summaries requires cross-verification, which demands considerable human and financial resources \citep{ahuja2021aspectnews, zhang2023macsum, chen2023unisumm}.
These potential obstacles collectively contribute to the scarcity of high-quality summary data.

From this consideration, instead of traditional naive fine-tuning on large amounts of training data, we would like to fully use the extremely limited amount of high-quality data to push the upper limit of PLMs' summarization ability.
To address this problem, we propose {\bf \alignsum}, a novel summarization preference alignment framework. 
First, we design a bottom-to-up data construction method {\bf Data Pyramid (DP)}, which consists of three components: extractive data, abstractive data, and human-annotated data. Different levels of data are collected with different methods.
% As illustrated in Table \ref{table:dp-type}, each type in the DP corresponds to a specific summarization approach. The volume of each data type decreases progressively, with acquisition difficulty increasing accordingly. Extractive data is derived directly from the original text using Gap Sentence Generation (GSG, \citet{zhang2020pegasus}), enhancing the model's ability to identify key information within documents. Abstractive data is generated by large language models (LLMs) such as LLaMA \citep{touvron2023llama}, which exhibit notable zero-shot summarization capabilities.
% The most crucial component is human-annotated data. Human preferences in summarization are highly variable, as exemplified by the saying ``\textit{there are a thousand Hamlets in a thousand people's eyes}''. Despite the impressive zero-shot capabilities of LLMs, their general summaries may not meet specific user preferences. To address this variability, we propose annotating a modest quantity of example data. This approach allows the model to better understand user intentions and align with human judgment.
DP is the core component of the alignment framework, after obtaining DP, we design the {\bf Gaussian Resampling} technique to smooth the length distribution of all summaries, and the two-stage {\bf Hierarchical Fine-Tuning (HFT)} to maximize the use of low-resource high-entropy human preference summary data.

We conduct experiments on human-annotated \textit{CNN/DailyMail} and \textit{BBC XSum} datasets proposed by \citet{wang2023element}, which reflects the implicit element-aware human writing preference.
We find that the pre-trained B{\footnotesize ART}-Large applied with AlignSum surpasses 175B GPT-3 on both automatic scores and human ratings, achieving amazing results in outperforming large models with small models and small amounts of preference data.
% , achieving single model SOTA results in automatic evaluations. Additionally, we recruited annotators to calculate the win rate of BART-Large (+DP) compared to 175B GPT3-CoT against carefully annotated human gold references. The results show that our method aligns more closely with human evaluations. 

% In summary, our contributions are:
% \begin{itemize}[leftmargin=10px]
%     \item We propose a novel fine-tuning data construction method named DP, comprising extractive data, abstractive data, and human-annotated data.
%     \vspace{-0.8em}
%     \item We introduce a novel AlignSum to align PLMs with human evaluations using a small amount of high-quality data.
%     \vspace{-0.8em}
%     \item We conduct extensive experiments and human evaluations on the Element-aware dataset to validate the effectiveness and generalization of our proposed framework.
% \end{itemize}

\section{\alignsum: Summarization Preference Alignment Framework}\label{dp_mothod}

We first formalize the summarization task: Given a document $D= d_1 d_2 ... d_n$ with length $n$, the goal is to generate a summary $S=s_1 s_2 ... s_m$ with length $m$, and usually $m \ll n$.
Our proposed preference alignment framework consists of three parts: Data Pyramid (DP), Gaussian Resampling, and Two-stage Hierarchical Fine-tuning (HFT). 

Figure \ref{fig:dp_overall} shows the overall framework:
Firstly, we construct the Data Pyramid using various methods such as extraction, LLM generation, and human annotation. 
Secondly, as the source data have different summary lengths, PLMs with this data would lead to inconsistent summary lengths. To address this issue, we utilize Gaussian Resampling to adjust the generated summary lengths to approximate the target length.
Finally, we apply a two-stage hierarchical fine-tuning strategy: initially training the PLMs on extractive and abstractive data to fit the general domain, followed by fine-tuning the justly fine-tuned PLMs on human-annotated data to align with human preference.
Details will be introduced in the following parts.

\subsection{Data Pyramid Construction}\label{sec:corpus}

\begin{table}[tbp]
\centering
\begin{tabular}{ccc}
\hline
\textbf{Style} / \textbf{Type} & \textbf{Difficulty} & \textbf{Volume} \\ \hline
Extractive       & Easy   & Large  \\
Abstractive     & Medium & Small  \\
Human-annotated & Hard   & Little \\ \hline
\end{tabular}
\caption{Features of summaries in Data Pyramid: summary style / type, acquisition difficulty, and data volume.}
\label{table:dp-type}
\end{table}

Data Pyramid comprises three levels: extractive, abstractive, and human-annotated data. From bottom to top, they are arranged in increasing quality and access difficulty, while the quantity decreases (as shown in Table \ref{table:dp-type}).
The first two are the two most generic styles in the summarization field, and we refer to them collectively as {\it generic data}; the last is the most critical part used to align human preferences, and we refer to it as {\it personalized data}.

\paragraph{Extractive Data.}
The extractive data constitutes the majority of the pre-training corpus and is the easiest to acquire.
We adopt the GSG technique proposed by \citet{zhang2020pegasus} to select the most important sentence as the pseudo summary $\hat{S}$:
\begin{equation}
    \begin{split}
        &\ \  r_i = \mathrm{Rouge} (d_i, D_{\setminus d_i}), \\
        &\ \  \hat{S} = \mathrm{argmax}_{d_i} \{r_i\}_{i=1}^n.
    \end{split}
\end{equation}
We use the R{\footnotesize OUGE}-1 metric \citep{lin2004rouge} to calculate the similarity and iterate through the entire document to find the most similar sentence as the pseudo summary. Unlike the method described by \citet{zhang2020pegasus}, we extract only a single sentence due to the variability in sentence lengths, as controlling by the number of sentences is unreliable. Instead, sample selection is based on the number of tokens in the Gaussian resampling stage.

\paragraph{Abstractive Data.}\label{dp_ad} 
The extractive data helps identify important sentences within a document but is insufficient for summarizing crucial information that spans multiple sentences. In contrast, LLMs are effective zero-shot summarizers, capable of extracting summary information across sentences and at the document level \citep{goyal2022news, zhang2023summit}. We use both system and user prompts to guide LLMs in summarizing the document $D$ and and generating the pseudo summary $\hat{S}$. 
As shown in Table \ref{zero-shot-prompt}, the system prompt specifies general requirements for accurate summarization. The document is then inserted before the user prompts, ensuring the LLM can read the entire document and adhere to user requirements. The user prompt is dataset-specific, setting the desired summary length and number of words.

\begin{table}[htbp]
\centering

\begin{tabular}{| p{0.12\textwidth} | p{0.32\textwidth} |}
\hline
Document &
  {\it E.g.}: Newcastle stand-in skipper Moussa Sissoko is facing disciplinary action after he was sent off following a reckless challenge on Liverpool midfielder Lucas ... \\ \hline
System Prompt &
  Generate a concise and coherent summary towards the given article and don't generate anything else. Make sure the summary is clear, informative, and well-structured. \\ \hline
Dataset-specific User Prompt &
  Summarize the article in \texttt{[sent num]} sentences around \texttt{[word num]} words. \\ \hline
\end{tabular}
\caption{Zero-shot Summarization prompt to generate Abstractive Data with LLM. }
\label{zero-shot-prompt}
\vspace{-0.2in}
\end{table}

\paragraph{Human-annotated Data.} Human-annotated data is the most critical component of DP for aligning with human preference. 
Training on data generated by adapted GSG and LLMs has allowed PLMs to acquire domain-specific knowledge. However, to generate summaries that align with human preferences, further fine-tuning on annotated data is necessary. This annotated data contains explicit user preferences and is easy to acquire without specific instructions, as PLMs can learn preferences through the data itself. To avoid the variability of random annotations, we use the Element-aware dataset provided by \citet{wang2023element}. This dataset adheres to specific instructions, incorporating both micro and macro demands (Details refer to Appendix \ref{element-aware-instruction}), ensuring consistent and high-quality human annotations.

\subsection{Gaussian Resampling}\label{gaussian-resample}
\begin{figure}[htbp]
    \centering
    \includegraphics[width=0.48\textwidth]{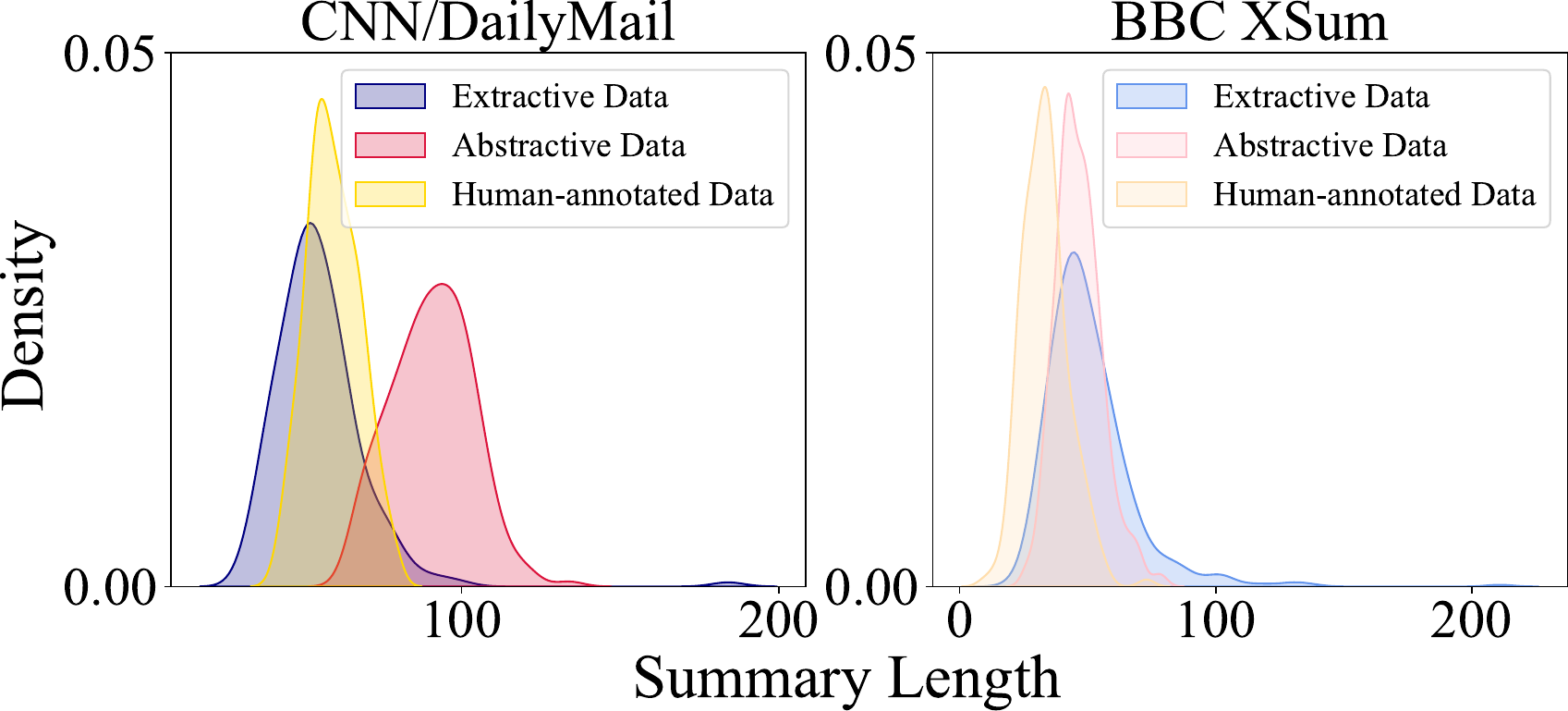}
    \vspace{-0.25in}
    \caption{Summary token length distributions of DP.}
    \label{fig:length}
    \vspace{-0.2in}
\end{figure}

DP draws from three distinct data sources, each with unique token length distributions for their pseudo summaries. 
As shown in Figure \ref{fig:length}, there are noticeable differences in summary token length distributions of extractive and abstraction data. 
Therefore, training directly with these disparate distributions can result in overly long or short summaries. 

To address this issue, we introduce the Gaussian Resampling technique to align all summary lengths with human-annotated summaries. Specifically, we model the token length distribution of human-annotated data as a Gaussian distribution:
\begin{equation}
     P(x) = \frac{1}{\sigma\sqrt{2\pi}} e^{-\frac{(x-\mu)^2}{2\sigma^2}},
\end{equation}
where $\mu$ and $\sigma$ represent the mean and standard deviation, respectively.
With a 95\% probability, the confidence interval for the token length distribution is [$\mu - 2\sigma$, $\mu + 2\sigma$].
We resample extractive and abstractive data within this interval to remove samples with excessively long or short pseudo summaries.

\subsection{Two-stage Hierarchical Fine-tuning}
\label{sec:hft}
Now we have obtained the resampled DP, a naive strategy is to fine-tune PLMs with them to enhance their summarization ability and align them with human preference simultaneously. However, this process can be challenging because the small amount of high-entropy data, which is crucial for alignment, can be interfered with by information from a large amount of low-entropy data \citep{wang2023retrodiff}, leading to the underutilization of DP.

To avoid this potential issue, we propose a two-stage hierarchical fine-tuning strategy.
Give a PLM $p_{\theta}$,
First is the {\bf generic fine-tuning stage}, where we fine-tune $p_{\theta}$ with the extractive and abstractive data to enhance its ability to generate domain-general summaries, obtaining a model $p_{\theta'}$.
Next is the {\bf personalized fine-tuning stage}, where we fine-tune $p_{\theta'}$ with the human-annotated data to create the final model $p_{\theta''}$ aligned with human preferences.

\paragraph{Why Hierarchical Fine-tuning?}
From a theoretical perspective of uncertainty reduction, we can explain the advantages of hierarchical fine-tuning using DP over hybrid fine-tuning.
We denote $X, Y, Z$ as the pre-trained data (intrinsic data of PLMs), generic data (extractive/abstractive data), and personalized data (human-annotated data), respectively. $p_{x;\theta}, p_{x,y;\theta}, p_{x,y,z;\theta}$ are models after pre-training, generic fine-tuning, and personalized fine-tuning, respectively. 
Let $J(p_{\theta})$ denote a random variable reflecting the summarization preference alignment ability of $p_{\theta}$, it is obviously that 
\begin{equation}
    \begin{aligned}
        &J(p_{\theta} = p_{x;\theta}) < J(p_{\theta} = p_{x,y;\theta}) \\
        &~~~~~~~~~~~~~~~~~~~~~ < J(p_{\theta} = p_{x,y,z;\theta}).
    \end{aligned}
\end{equation}

In general, generic data enhances the performance of downstream tasks, whereas task-specific data compromises the generalized capabilities of the model, \emph{i.e.}, the ``Alignment Tax'' \citep{ouyang2022training,dong2023raft}.
Therefore, we can intuitively make the following assumptions about the relationship between alignment uncertainty and alignment ability $J(p_{\theta})$ of model $p_{\theta}$:

\newtheorem{assumption}{Assumption}[section]
% \begin{assumption}
% \label{assumption1}
% For any data $S \in \{X, Y, Z\}$, its alignment uncertainty is lower when conditioned on a model with stronger alignment ability, i.e., 
% \begin{equation}
%     \begin{aligned}
%         &H(S|J(p_{\theta} = p_{x;\theta})) 
%         > H(S|J(p_{\theta} = p_{x,y;\theta})) \\
%         >& H(S|J(p_{\theta} = p_{x,y,z;\theta})) > 0
%     \end{aligned}
% \end{equation}
% \end{assumption}

\begin{assumption}
\label{assumption2}
For hierarchical data $\{X, Y, Z\}$, data at the lower level of DP enhances the model's ability of the upper-level tasks, but data at the upper level impairs the model's ability of the lower-level tasks, i.e.,
\begin{equation}
    \begin{aligned}
        &1. ~H(Z|J(p_{\theta} = p_{x;\theta})) 
        > H(Z|J(p_{\theta} = p_{x,y;\theta})) \\
        &~~~~~~~~~~~~~~~~~~~~~~~~~~~~~~~~> H(Z|J(p_{\theta} = p_{x,y,z;\theta})), \\
        &2. ~H(Y|J(p_{\theta} = p_{x;\theta})) 
        > H(Y|J(p_{\theta} = p_{x,y;\theta})), \\
        &3. ~H(Y|J(p_{\theta} = p_{x,y;\theta})) < H(Y|J(p_{\theta} = p_{x,y,z;\theta})) \\
        &4. ~H(X|J(p_{\theta} = p_{x;\theta})) 
        < H(X|J(p_{\theta} = p_{x,y;\theta})) \\
        &~~~~~~~~~~~~~~~~~~~~~~~~~~~~~~~~~< H(X|J(p_{\theta} = p_{x,y,z;\theta})). \\
    \end{aligned}
\end{equation}
\end{assumption}

We derive the uncertainty reductions before and after fine-tuning for both fine-tuning strategies:
\begin{itemize}[leftmargin=10px]
    \item hybrid fine-tuning: \begin{equation}
        \begin{aligned}
            G_{hy} = &\left| H(Y,Z|J(p_{\theta} = p_{x,y,z;\theta})) \right.
            \\&- \left. H(Y,Z|J(p_{\theta} = p_{x;\theta})) \right|
        \end{aligned}
    \end{equation}
    \item hierarchical fine-tuning: 
    \begin{small}
    \begin{equation}
        \begin{aligned}
            G_{hi} = &| \underbrace{H(Y|J(p_{\theta} = p_{x,y;\theta})) - H(Y|J(p_{\theta} = p_{x;\theta}))}_{\text{generic fine-tuning stage}} | \\
            +&| \underbrace{H(Z|J(p_{\theta} = p_{x,y,z;\theta})) - H(Z|J(p_{\theta} = p_{x,y;\theta}))}_{\text{personalized fine-tuning stage}} |
        \end{aligned}
    \end{equation}
    \end{small}
\end{itemize}

We can prove that  $G_{hi} > G_{hy}$ holds constant for any model $p_{\theta}$ and data sets $X, Y, Z$ under the Assumption \ref{assumption2}. This means the uncertainty reduction from hierarchical fine-tuning is greater, leading to a better alignment performance. Appendix \ref{sec:theory} shows the complete proof.
Table \ref{tab:stage_training_and_resample} in Section \ref{sec:alignsum_ablation} also demonstrates the need for hierarchical fine-tuning from an empirical perspective.

\begin{table*}[]
\centering
\resizebox{\textwidth}{!}{
    \begin{tabular}{@{}l|cccc|cccc@{}}
    \toprule
    \multirow{2}{*}{\diagbox[height=27pt,trim=l]{Model}{Dataset}} & \multicolumn{4}{c|}{ \textbf{ \textit{CNN/DailyMail}}}                                                                     & \multicolumn{4}{c}{\textbf{ \textit{BBC XSum}}} \\ \cmidrule(l){2-9} 
                       & \multicolumn{1}{c}{\textbf{R{\footnotesize OUGE}-1}} & \multicolumn{1}{c}{\textbf{R{\footnotesize OUGE}-2}} & \multicolumn{1}{c|}{\textbf{R{\footnotesize OUGE}-L}} & \textbf{BERTScore} & \multicolumn{1}{c}{\textbf{R{\footnotesize OUGE}-1}} & \multicolumn{1}{c}{\textbf{R{\footnotesize OUGE}-2}} & \multicolumn{1}{c|}{\textbf{R{\footnotesize OUGE}-L}} & \textbf{BERTScore} \\ \midrule

    \multicolumn{9}{c}{ \textbf{Direct Generation (w/ LLMs)}} \\
    \midrule
    \textit{175B GPT-3, 0-shot}                  & {\it 42.98}                        & {\it 19.48}                        & \multicolumn{1}{c|}{\it 28.33}   & {\it 0.8943}    & {\it 38.50}                        & {\it 15.09}                        & \multicolumn{1}{c|}{\it 29.09}   & \textbf{\it 0.8981}    \\
    {\it ~~~~~~ w/SumCoT, 0-shot}            & {\it 49.73}                        & {\it 26.10}                        & \multicolumn{1}{c|}{\it 36.29}   & {\it 0.9080}    & {\it 44.36}                        & {\it 19.93}                        & \multicolumn{1}{c|}{\it 34.70}   & {\it 0.9053}    \\ 
    
    \textit{GPT-3.5-Turbo, 0-shot}                  & {\it 41.82}                        & {\it 18.50}                        & \multicolumn{1}{c|}{\it 27.63}   & {\it 0.8958}    & {\it 31.38}                        & {\it 13.37}                        & \multicolumn{1}{c|}{\it 23.05}   & \textbf{\it 0.8865}    \\
    \textit{~~~~~~ w/Style, 0-shot}                  & {\it 45.62}                        & {\it 19.51}                        & \multicolumn{1}{c|}{\it 31.52}   & {\it 0.8997}    & {\it 41.80}                        & {\it 18.31}                        & \multicolumn{1}{c|}{\it 31.58}   & \textbf{\it 0.8984}    \\
    \textit{~~~~~~ w/Style, 1-shot}                  & {\it 45.71}                        & {\it 18.70}                        & \multicolumn{1}{c|}{\it 29.98}   & {\it 0.8996}    & {\it 41.32}                        & {\it 17.19}                        & \multicolumn{1}{c|}{\it 31.52}   & \textbf{\it 0.8985}    \\

    \textit{L{\footnotesize LaMA}-2-7B}                & {\it 44.78}                        & {\it 18.83}                        & \multicolumn{1}{c|}{\it 29.65}   & {\it 0.8985}    & {\it 37.99}                        & {\it 14.20}                        & \multicolumn{1}{c|}{\it 28.72}   & {\it 0.8952}    \\
    {\it L{\footnotesize LaMA}-3-8B}  & {\it 46.27}                        & {\it 20.23}                        & \multicolumn{1}{c|}{\it 31.23}   & {\it 0.9011}    & {\it 40.34}                        & {\it 16.12}                        & \multicolumn{1}{c|}{\textit{30.00}}   & {\it 0.8959}  \\
    
    \midrule

    \multicolumn{9}{c}{ \textbf{Naive Fine-tuning (w/ PLMs)}} \\
    \midrule
    B{\footnotesize ART}-Base                   & 44.67                        & 20.43                        & \multicolumn{1}{c|}{29.86}   & 0.8754    & \underline{30.04}                        & 8.95                        & \multicolumn{1}{c|}{21.71}   & 0.8787    \\
    B{\footnotesize ART}-Large                   & 46.01                        & 21.92                        & \multicolumn{1}{c|}{\underline{32.08}}   & 0.8851    & 28.73                        & 8.80                        & \multicolumn{1}{c|}{20.96}   & 0.8811    \\
    T5-Large                   & 43.64                       & 19.23                        & \multicolumn{1}{c|}{30.76}   & 0.8842    & 29.83                        & \underline{9.14}                        & \multicolumn{1}{c|}{\underline{21.99}}   & 0.8790    \\
    P{\footnotesize EGASUS}                   & 41.39                        & 15.66                        & \multicolumn{1}{c|}{27.26}   & 0.8706    & 29.26                        & 7.56                        & \multicolumn{1}{c|}{21.26}   & \underline{0.8825}    \\
    B{\footnotesize RIO}  & \underline{46.66} & \underline{22.35}   & \multicolumn{1}{c|}{31.01} & \underline{0.8876}  & 28.45                        & 8.34                        & \multicolumn{1}{c|}{21.05}   & 0.8787 \\
    
    \midrule
    
    % Our             & 41.24                        & 16.27                        & \multicolumn{1}{c|}{26.90}   & 0.8815    & 35.73                        & 13.45                        & \multicolumn{1}{c|}{25.97}   & 0.8815    \\ \midrule

    \multicolumn{9}{c}{ \textbf{\alignsum ~(w/ PLMs, Ours)}} \\
    \midrule
    
    L{\footnotesize LaMA}-2-7B (w/ HD)               & 44.37                        & 18.17                        & \multicolumn{1}{c|}{28.96}   & 0.8906    & 37.08                        & 14.07                        & \multicolumn{1}{c|}{28.57}   & 0.8937    \\
     B{\footnotesize ART}-Large (w/ HD)                & 46.57                        & 21.97                        & \multicolumn{1}{c|}{32.00}   & 0.9040    & 40.19                        & 14.95                        & \multicolumn{1}{c|}{28.74}   & 0.8915    \\
     B{\footnotesize ART}-Base (w/ full DP)                & 45.01                        & 20.51                        & \multicolumn{1}{c|}{31.79}   & 0.8998    & 39.88                        & 16.46                        & \multicolumn{1}{c|}{30.45}   & 0.8911    \\
    B{\footnotesize ART}-Large (w/ full DP)                 & \textbf{48.83}                        & \textbf{24.11}                        & \multicolumn{1}{c|}{\textbf{34.16}}   & \textbf{0.9058}    & \textbf{42.38}                        & \textbf{17.75}                        & \multicolumn{1}{c|}{\textbf{31.64}}   & {\bf 0.8962}    \\ \bottomrule
    \end{tabular}
}
\caption{\label{main-result}Automatic metrics R{\footnotesize OUGE}-1/2/L and B{\footnotesize ERT}Score Performances of LLMs and PLMs under naive fine-tuning and our \alignsum ~settings on human preference Element-Aware dataset \citep{wang2023element}.
\textit{Italic} means LLM results inferred via API or pre-trained weights, details are shown in Appendix \ref{llm-inference}.
\textbf{Bold} represents the best performances among all fine-tuned models, ``w/ style'' means style control with prompt in Table \ref{zero-shot-prompt},  ``w/ HD'' indicates fine-tuning with HD data, and ``w/ full DP'' represents our final model. The result of B{\footnotesize ART} (w/ HD) is sampled 5 times and reports the mean. Details are shown in Appendix \ref{random_exp}.}
\vspace{-0.18in}
\end{table*}

\section{Experiments}

\subsection{Setup}

\paragraph{Dataset.}
We conduct DP construction and experiments on two extensively used news datasets, \textit{CNN/DailyMail} \citep{nallapati2016abstractive} and \textit{BBC XSum} \citep{narayan2018don}. 
For generation of extractive data (ED) and abstractive data (AD), we divide the standard training set with an 8:2 ratio to generate ED and AD for training, respectively.
For human-annotated data (HD) that implicitly reflect human preference\footnote{These preferences reflect professional implicit writing styles embedded in texts and are difficult to capture explicitly.}, we adopt the Element-Aware \textit{CNN/DailyMail} and \textit{BBC XSum}, which is the high-quality rewritten version \citep{wang2023element} of the two datasets (each 200 samples).
Refer to Appendix \ref{element-aware-instruction} for detailed preference features, and Appendix \ref{data_case} for data examples of the two datasets.
For testing, we randomly split HD into a training set and a test set, each containing 100 samples.

\paragraph{Data Statistics.}
Table \ref{dataset-statistic} shows the total count and token length distribution of pseudo summary in DP. The training set and test set are randomly sampled from the Element-Aware dataset. ED extracts the most important sentence from the original document, and the token length varies greatly. After the Gaussian Resampling, the ED$_r$ standard deviation slows down. Although the mean length of ED$_r$ is smaller than the HD, it all falls into the HD's distribution confidence interval. The same for AD$_r$, standard deviation slows down and all data token lengths fall into the desired range.

% \usepackage{booktabs}
% \usepackage{multirow}
% \usepackage{graphicx}

% \usepackage{booktabs}

\paragraph{Baselines.}
We choose two settings for baselines: (i) Zero-Shot Generation with LLMs, we select 175B GPT-3 \citep{brown2020language} and GPT-3.5-Turbo; (ii) Naive Fine-tuning with PLMs, means directly fine-tuning models with standard training sets of corresponding datasets. We select B{\footnotesize ART-}Large, B{\footnotesize ART-}Base \citep{lewis2019bart}, T5-Large \citep{raffel2020exploring}, P{\footnotesize EGASUS} \citep{zhang2020pegasus}, B{\footnotesize RIO} \citep{liu2022brio}, L{\footnotesize LaMA}-2-7B \citep{touvron2023llama}, and L{\footnotesize LaMA}-3-8B \citep{llama3}.
All model weights are downloaded from HuggingFace.
Refer to Appendix \ref{appendix:packages} for more details.
% In the subsequent analysis, we mainly compare our method under PLM settings for fairness.

\paragraph{Implementation.}
We use the pre-trained B{\footnotesize ART}-Large for the backbone of \alignsum ~and L{\footnotesize LaMA}-2-7B for generating abstractive data due to its ease of use, with Appendix \ref{appendix:llm_ad} showing that different LLMs perform similarly.
For PLMs, we truncate documents to 1024 tokens and target summaries to 128 tokens following \citet{zhang2020pegasus}. Given that LLMs can handle up to 4096 tokens, we truncate the original documents to 2048 tokens for LLM inference. 
To ensure a fair comparison, we fine-tune all PLMs with both extractive and abstractive data for $3$ epochs, using a learning rate of $5e^{-5}$ and a batch size of $128$. Due to the limited amount (only $100$ samples) of human-annotated data, we fine-tune them with 20 epochs, keeping the other hyperparameters unchanged.

\begin{table}[t]
\centering
\begin{tabular}{@{}l|cc@{}}
\toprule
\multirow{2}{*}{\textbf{Data}}     & \multicolumn{1}{c|}{\textbf{\textit{CNN/DailyMail}}} & \textbf{\textit{BBC XSum}}          \\ \cmidrule(l){2-3} 
                             & \multicolumn{2}{c}{\textbf{Sample Number\ /\ Length Mean{\tiny $\pm$std}}}                       \\

\midrule
\multicolumn{3}{c}{\bf Training Set} \\
\midrule

\textcolor{lightgray}{ED}  & \textcolor{lightgray}{229k\ / \ 55{\tiny $\pm$17}}      &  \textcolor{lightgray}{163k\ /\ 51{\tiny $\pm$53}}   \\
ED$_r$    & 224k\ / \ 54{\tiny $\pm$12}                  &   107k\ /\ 41{\tiny $\pm$7}     \\
\textcolor{lightgray}{AD} & \textcolor{lightgray}{57k\ / \ 91{\tiny $\pm$13}}        & \textcolor{lightgray}{41k\ /\ 46{\tiny $\pm$9}}  \\
AD$_r$ & 40k\ / \ 85{\tiny $\pm$8}                    & 32k\ / \ 43{\tiny $\pm$6}    \\ 
HD  & 0.1k\ / \ 64{\tiny $\pm$17}                    & 0.1k\ /\ 34{\tiny $\pm$10}     \\
\midrule
\multicolumn{3}{c}{\bf Test Set} \\
\midrule
HD          & 0.1k\ / \ 66{\tiny $\pm$15}                     & 0.1k\ /\ 33{\tiny $\pm$8}      \\
\bottomrule
\end{tabular}%
\caption{\label{dataset-statistic}Sample numbers and pseudo summary token length statistics. We use B{\footnotesize ART}-Large as the tokenizer. ED$_r$ and AD$_r$ mean ED and AD after Gaussian Resampling, respectively. Data colored by \textcolor{lightgray}{gray} are not involved in the actual training process.}
\vspace{-0.2in}
\end{table}

\subsection{Automatic Evaluation}

Automatic evaluation usually contradicts human evaluation when referenced gold summaries are low-quality \citep{goyal2022news}. However, when references are high-quality, automatic evaluation results are more consistent with that of human evaluation as verified by \citet{wang2023element}.

Table \ref{main-result} presents the overall results:

\vspace{-0.05in}
\paragraph{Comparisons with Naive Fine-tuned PLMs.}
Compared with SOTA results of PLMs under the naive fine-tuning setting, B{\footnotesize ART}-Large with \alignsum ~improves R{\footnotesize OUGE}-1/2/L by over +2.17/+1.76/+2.08
points on \textit{CNN/DailyMail} and by +12.37/+8.61/+9.65 points on \textit{BBC XSum}, even though these models are pre-trained with the original low-quality dataset. BERTScore for B{\footnotesize ART}-Large (w/ full DP) is also higher than for the other PLMs. These results indicate that: 
(1) Fine-tuning on low-quality original datasets does not enhance human alignment; 
(2) Further fine-tuning on HD data significantly boosts performance, as seen with B{\footnotesize ART}-Large (w/ HD) improving B{\footnotesize ART}-Large (w/ Naive Fine-tuning) by nearly +0.5 points and +12 points on \textit{CNN/DailyMail} and \textit{BBC XSum}.

\vspace{-0.05in}
\paragraph{Comparisons with Zero-shot LLMs.}
Compared to LLMs with the zero-shot setting, since summarization is unsuitable for few-shot due to restricted context, we find that even though part of DP is generated from L{\footnotesize LaMA}-2-7B, its R{\footnotesize OUGE} and BERTScore are lower than B{\footnotesize ART}-Large (w/ full DP). 
Additionally, GPT-3 performs worse than L{\footnotesize LaMA}-2-7B because we control the generation length in Section \ref{dp_ad}, whereas GPT-3 is only prompted with ``\textit{Summarize the above article}'' as used in \citep{goyal2022news, sanh2021multitask}. B{\footnotesize ART}-Large (w/ full DP) is slightly worse than GPT-3 (w/CoT), which is expected since GPT-3 was carefully prompted according to the data annotation protocol, making it less adaptable to other writing styles. In contrast, our model aligns with specific human preferences using modest HD data.

% \begin{itemize}[leftmargin=0.3cm]
%     \item Fine-tuning PLMs with human-annonated significantly enhances performance. In contrast, fine-tuning LLMs with a small amount of human-annonated data proves less effective.
%     \item The performance of BART-Large(DP) surpasses all other models and is marginally inferior to the 175B GPT-3 employing SumCoT \citep{wang2023element}. However, the prompt-based SumCoT aligns closely with the data annotation strategy, limiting its scalability. Conversely, BART-Large(DP) learns the summarization model directly from data, offering an end-to-end solution.
% \end{itemize}

\vspace{-0.05in}
\paragraph{Comparisons with Fine-tuned LLMs.}
Compared to fine-tuning LLMs, we use LoRA \citep{hu2021lora} to fine-tune LLaMA2-7B. Despite LoRA having significantly fewer trainable parameters than BART fine-tuning, its memory consumption during training exceeds that of BART, even with a batch size of 1. This makes training unfeasible on consumer-grade hardware. Additionally, fine-tuning with only 100 HD samples fails to improve performance and may even decrease it, as shown in Table \ref{main-result}. This is because high-quality fine-tuning typically requires datasets on the order of tens of thousands \citep{deng2023k2,zhao2024chemdfm}. Furthermore, this fine-tuning process may negatively impact the LLMs' other capabilities, such as mathematical and logical reasoning.

\vspace{-0.05in}
\paragraph{Significance Test.}
Given the limited sample size of 100 for each dataset, we apply Analysis of Variance (ANOVA, ~\citet{st1989analysis}) to determine if statistically significant differences exist between the random experiments. Table \ref{tab:rouge_scores_significance} indicates significant differences, as nearly all p-values are below 0.05 ($p < 0.05$).

\begin{table}[ht]
\centering
\resizebox{1\columnwidth}{!}{
\begin{tabular}{lccc}
\toprule
\textbf{Dataset} & \textbf{R{\footnotesize OUGE}-1} & \textbf{R{\footnotesize OUGE}-2} & \textbf{R{\footnotesize OUGE}-L} \\ \midrule
\textit{CNN/DailyMail} & 0.013 & 2.59e-6 & 5.89e-4 \\ 
\textit{BBC XSum} & 0.022 & 0.056 & 0.028 \\ \bottomrule
\end{tabular}
}
\caption{R{\footnotesize OUGE}-1/2/L p-values of multiple experiments on \textit{CNN/DailyMail} and \textit{BBC XSum}.}
\label{tab:rouge_scores_significance}
\end{table}

\vspace{-0.05in}
\subsection{Human Evaluation}

We conduct human evaluations to compare the performances of PLMs with \alignsum ~and 175B GPT-3 (w/CoT) for it is the strongest LLM in automatic evaluation. Typically, human evaluation is reference-free \footnote{To evaluate \alignsum ~more comprehensively, we also conduct a reference-free human evaluation, with details shown in Appendix \ref{appendix:reference-free-human}. } and involves in informativeness, conciseness, readability, and faithfulness \citep{bao2023gemini,liu2023gpteval}. 
We instead use a reference-based evaluation for two reasons: (1) The Element-Aware dataset has included expert-written high-quality references; (2) Referenced summaries represent a specific writing style, and evaluating only the four qualities would overlook implicit preference features captured by HD.

\begin{figure}[!t]
    \centering
    \includegraphics[width=0.48\textwidth]{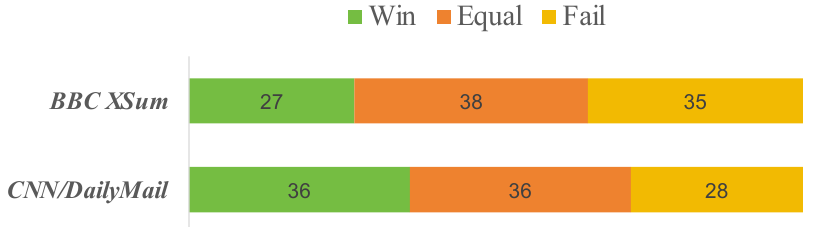}
    \caption{Reference-based human evaluation of BART (w/ full DP) and GPT-3 (w/CoT) compared to the golden reference on \textit{CNN/DailyMail} and \textit{BBC XSum}.}
    \label{fig:human_eval}
\vspace{-0.2in}
\end{figure}

Given generated summaries of B{\footnotesize ART}-Large with \alignsum ~and 175B GPT-3 (w/CoT), and expert-written high-quality reference summaries, human evaluation follows these instructions:

\begin{itemize}[leftmargin=10px]
% \vspace{-0.8em}
    \item {\it Length Pre-screening}: Summaries that are too long or short compared to the reference text are considered ``Fail''. If both generated summaries ``Fail'', they are considered ``Equal''.
    % \vspace{-0.4em}
    \item {\it Overall Evaluation}: If generated summaries have similar lengths, we compare their informativeness. Informativeness is defined by characteristic elements: entities, dates, events, and results \citep{wang2023element}, each denoted as a set ($S_{\rm en}$, $S_{\rm da}$, $S_{\rm ev}$, $S_{\rm re}$) for a summary $S$. Let generated summaries of \alignsum ~and 175B GPT-3 (w/CoT) as $\hat{S}_1$ and $\hat{S}_2$ and the gold reference as $G$, we can define informativeness for each generated summary:
    \begin{equation}
        \begin{split}
            \mathrm{Info}_i &= \sum_j |G_{j} \cap [\hat{S}_i]_{j}|, \ i=1,2 \\
            j &\in \{\rm en, da, ev, re\}, 
        \end{split}
    \end{equation}
    where $|\cdot|$ represents the number of elements in the set. If Info$_1$ $>$ Info$_2$, then \alignsum ~ ``Win''; if Info$_1$ $=$ Info$_2$, they are ``Equal''; otherwise, \alignsum ~``Fail''.
\end{itemize}

We recruited one Ph.D. student and two Master students majoring in Computer Science to conduct the evaluation following the above instructions. The majority vote is selected as the final rating, and the human evaluation results are presented in Figure \ref{fig:human_eval}. Although B{\footnotesize ART}-Large with \alignsum ~slightly underperforms in automatic evaluation compared to GPT-3 (w/CoT), it achieves ``Win'' and ``Equal'' ratings of up to 65\% and 72\% on the \textit{BBC XSum} and \textit{CNN/DailyMail} datasets, respectively. This demonstrates that {\bf our \alignsum ~can effectively align PLMs with human evaluation standards without requiring billions of model parameters or sophisticated prompt designs}. Additional case studies are provided in the Appendix \ref{case_study}.

\section{Ablation Study}\label{ablation}

\subsection{Components of \alignsum}
\label{sec:alignsum_ablation}

We conduct ablation for \alignsum's components to verify their effectiveness. Table \ref{tab:stage_training_and_resample} shows the results under different component combinations.

\begin{table}[t]
  \centering
    %   \vspace{-0.15in}
    \resizebox{1\columnwidth}{!}{
    \begin{tabular}{ccc|ccc}
    
    \toprule
    \multicolumn{3}{c|}{\textbf{Component}}  & \multicolumn{3}{c}{\multirow{1}{*}{\textbf{Metric}}}\\
    \midrule
    DP & GR & HFT & \textbf{R{\footnotesize OUGE}-1} & \textbf{R{\footnotesize OUGE}-2} & \textbf{R{\footnotesize OUGE}-L} \\
    \midrule

    \multicolumn{6}{c}{\textit{\textbf{CNN/DailyMail}}} \\
    \midrule
     &  &  & 41.38 & 18.35 &  26.05 \\
    \checkmark  &  &  & 39.01 & 14.59 & 25.83 \\
    \checkmark  & \checkmark & & 37.63 & 13.86 & 24.75 \\
    \checkmark  &  & \checkmark & 47.53 & 21.78 & 32.56 \\
     \rowcolor{gray!20} 
     \checkmark & \checkmark & \checkmark & \textbf{48.39}       & \textbf{23.33}       & \textbf{34.47} \\
    \midrule

    \multicolumn{6}{c}{\textbf{\textit{BBC XSum}}} \\
    \midrule
     &  &  & 34.86 & 12.22 & 24.19 \\
    \checkmark  &  &  & 38.46 & 16.79 & 28.68 \\
    \checkmark  & \checkmark & & 39.71 & 16.92 & 28.51 \\
    \checkmark  &  & \checkmark & \textbf{44.58} & 19.60 & \textbf{32.90} \\
     \rowcolor{gray!20} 
     \checkmark & \checkmark & \checkmark & 43.68 & \textbf{19.73} & 32.15 \\
    \bottomrule
    \end{tabular}%
    }
% \vspace{-0.1in}
\caption{Ablation study on the effectiveness enhancement from different components of \alignsum, including Data Pyramid (DP), Gaussian Resampling (GR), and Hierarchical Fine-Tuning (HFT).}
\label{tab:stage_training_and_resample}
\vspace{-0.2in}
\end{table}%

\paragraph{Gaussian Resampling.}
When adding the Gaussian resampling component, the performance on \textit{CNN/DailyMail} improves, where ``DP+GR+HFT'' improves upon ``DP+HFT'' by +0.86/+1.55/+1.91 points in R{\footnotesize OUGE}-1/2/L, respectively. 
However, when on \textit{BBC XSum}, we observe a slight performance degradation. This may be attributed to the raw data distribution closely matching the target distribution, while the Gaussian Resampling filters out nearly 20-30\% of the raw data.
% Although DP-R shows little to no improvement over DP, the average summary length trained with the Gaussian resample is closer to the ground reference.

\paragraph{Hierarchical Fine-Tuning.}
When adding the Hierarchical Fine-Tuning component, we observe substantial improvements in both datasets, with an average of +10 points improvement on \textit{CNN/DailyMail} and +4 points improvement on \textit{BBC XSum}.
This validates the conclusion proved in Section \ref{sec:hft} that the two-stage fine-tuning helps to reduce the information loss of high-entropy variables and maximize the use of the limited preference summary data, and also demonstrates that mixed fine-tuning tends to dilute the impact of HD within the larger volumes of ED and AD.

\subsection{Components of Data Pyramid}

\begin{figure}[t]
    \centering
    \includegraphics[width=0.48\textwidth]{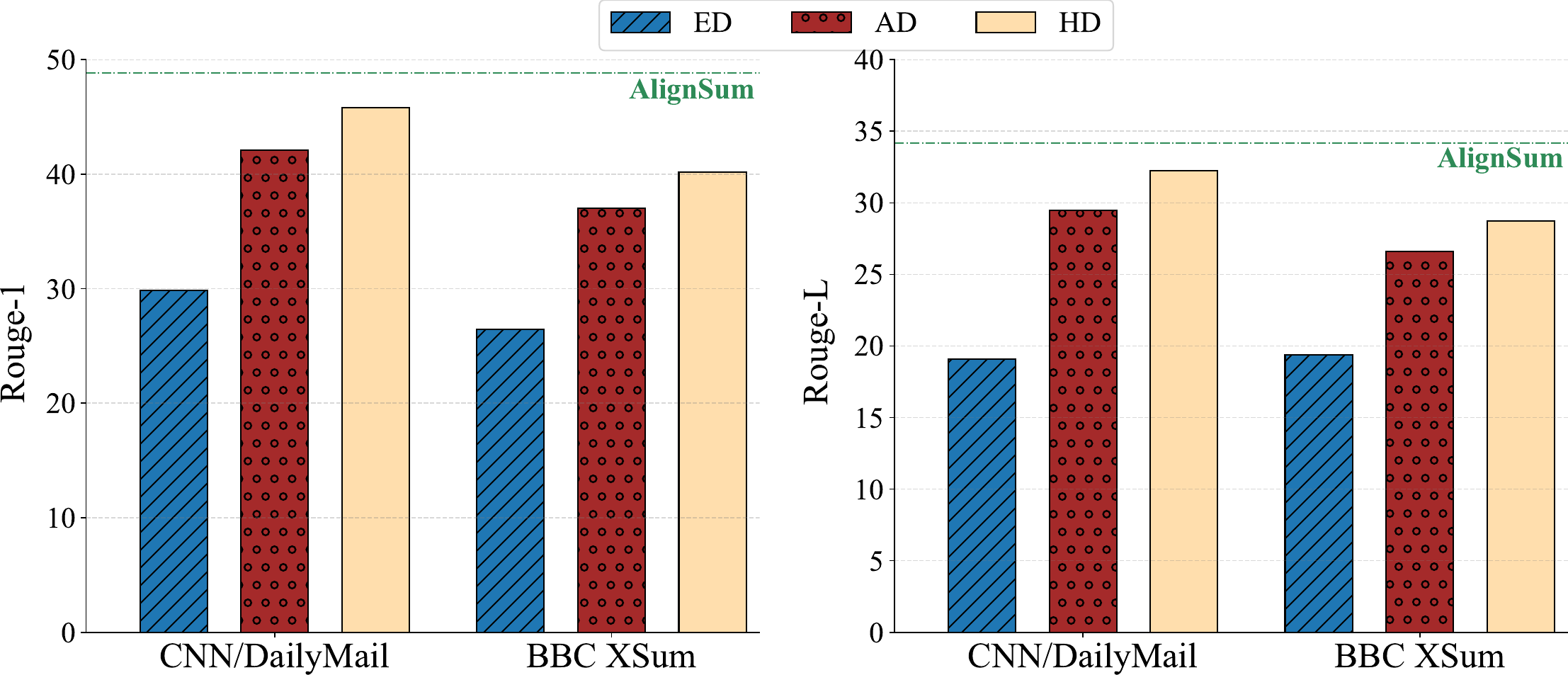}
    \caption{R{\footnotesize OUGE}-1/L of fine-tuning B{\footnotesize ART}-Large with ED, AD, HD on \textit{CNN/DailyMail} and \textit{BBC XSum}.}
    \label{fig:componets_aba}
% \vspace{-0.2in}
\end{figure}

We fine-tune B{\footnotesize ART}-Large with ED, AD, and HD separately. Figure \ref{fig:componets_aba} shows the R{\footnotesize OUGE}-1/L results on the two datasets. The importance of high-quality data becomes increasingly evident, as fine-tuning with any single data type cannot outperform our proposed framework with DP.

\subsection{Human-annotated Data (HD) Size}
Table \ref{few-shot} illustrates the impact of varying amounts of human-annotated data on B{\footnotesize ART}'s ability to learn user summary patterns. With 50 training samples, B{\footnotesize ART}'s performance already surpasses that of all pre-trained models and LLMs. Furthermore, as the amount of human-annotated data increases, the model's performance improves correspondingly.

\begin{table}[t]
\resizebox{\columnwidth}{!}{%
\begin{tabular}{@{}c|cccc@{}}
\toprule
HD size & \multicolumn{1}{c}{R{\footnotesize OUGE}-1} & \multicolumn{1}{c}{R{\footnotesize OUGE}-2} & \multicolumn{1}{c}{R{\footnotesize OUGE}-L} & BERTScore \\ \midrule
10  & 44.72 & 18.96 & 29.48 & 0.8855 \\
50  & 47.38 & 21.67 & 31.73 & 0.8897 \\
100 & 48.04 & 22.67 & 33.38 & 0.9050 \\ \bottomrule
\end{tabular}%
}
\caption{\label{few-shot}R{\footnotesize OUGE}-1/2/L and BERTScore results on \textit{CNN/DailyMail} under various HD sizes.}
\vspace{-0.2in}
\end{table}

\subsection{Abstractive Data (AD) Scaling} 
High-quality HD is difficult to acquire, but AD is also useful and relatively easier to generate.
% , and we used only 20\% of the raw dataset to generate AD in DP. 
% In our experimental setting, we select only 20\% documents of the raw dataset to generate AD in DP. 
% Figure \ref{fig:ad-scaling} shows the ROUGE-1 scores of $p_{\theta'}$ and  $p_{\theta''}$ on \textit{BBC XSum} with varying proportions of AD and ED. Although the ROUGE-1 score grows with higher proportions of AD after generic fine-tuning, different models fluctuate slightly with personalized fine-tuning. We adopt the DP rather than training with more AD or total AD for two reasons: 1) Generating AD is slightly more complex than extracting ED. 2) More importantly, DP introduces greater diversity. While ED helps identify key sentences, AD struggles to handle this effectively.
Table \ref{tab:ad_scaling} presents the results of scaling AD on \textit{BBC XSum}.
When keeping the training sample size fixed while varying the proportion of AD and ED, it is evident that increasing the amount of AD does not enhance performance. In fact, ED plays a critical role in improving model performance, as performance significantly degrades when $\tau = 1:0$, which indicates exclusive use of AD.
In this setting, increasing AD would lead to a reduction in ED, introducing multiple variable changes. 
Then, we fix the ED component and utilize the entire training data to generate them, while we use different proporation of training data to generate AD again, which would increase the total training sample size. However, the results again demonstrate that increasing AD does not improve performance, due to a mismatch between the data distribution of AD generated by large LLMs and the test set.
In contrst, DP enhances performance by introducing greater diversity, as ED effectively identifies key sentences, while AD struggles to do so.

% \begin{figure}[t]
%     \centering
%     \includegraphics[width=0.44\textwidth]{figures/ad-scaling.pdf}
%     \caption{ROUGE-1 scores for fine-tuning BART with AD and ED generated by different proportions of original \textit{BBC XSum}. The horizontal axis represents the AD ratio, "100\%" means all data are used to generate AD.}
%     \label{fig:ad-scaling}
% \vspace{-0.2in}
% \end{figure}

\begin{table}[t]
  \centering
    %   \vspace{-0.15in}
    % \resizebox{1\columnwidth}{!}{
    \begin{tabular}{c|ccc}
    
    \toprule

    \multicolumn{4}{c}{\textit{\textbf{Fixed Sample Pool}}} \\
    \midrule
    $\tau$  &\textbf{R{\footnotesize OUGE}-1} & \textbf{R{\footnotesize OUGE}-2} & \textbf{R{\footnotesize OUGE}-L} \\
    \midrule
    0.2:0.8 & 43.69 & 19.73 & 32.15 \\ 
    0.5:0.5 & 44.20 & \textbf{21.14} & \textbf{33.88} \\ 
    0.8:0.2 & 43.00 & 19.43 & 33.11 \\ 
    1:0 & 44.13 & 19.50 & 32.93 \\ 
    \midrule
    
    \multicolumn{4}{c}{\textit{\textbf{Increasing Sample Pool}}} \\
    \midrule
    
     $\sigma$  &\textbf{R{\footnotesize OUGE}-1} & \textbf{R{\footnotesize OUGE}-2} & \textbf{R{\footnotesize OUGE}-L} \\
    \midrule
    0.2 & 44.38 & 19.89 & 33.43 \\ 
    0.5 & 43.10 & 19.48 & 32.46 \\ 
    0.8 & \textbf{44.56} & 20.20 & 33.29 \\ 
    1 & 43.56 & 18.84 & 32.64 \\ 

    \bottomrule
    \end{tabular}%
    % }
% \vspace{-0.1in}
\caption{Results of scaling abstractive data (AD) on \textit{BBC XSum}. \textit{\textbf{“Fixed Sample Pool”}} refers to maintaining the original training set, where $\tau$ denotes the proportion used to generate AD and ED. In contrast, \textit{\textbf{“Increasing Sample Pool”}} indicates that the original training data is utilized twice: initially for generating ED, with $\sigma$ representing the proportion used to generate AD.}
\label{tab:ad_scaling}
\vspace{-0.2in}
\end{table}%

\section{Related Work}

\paragraph{Extractive Summarization.} 
Extractive summarization aims to extract sentences from given documents \citep{zhong2020extractive}. Current approaches formulate this task as a classification or matching problem using recurrent neural networks \citep{cheng2016neural, nallapati2016abstractive}, pre-trained language models \citep{liu2019text,wang2022noise}, large language models \citep{zhang2023extractive}, and even diffusion models \citep{zhang2023diffusum}. Although extractive summarization cannot effectively synthesize summary information across sentences at the document level, they are always grammatically correct and faithful to the original text. Therefore, we utilize extractive summarization as the basis to help identify key sentences in the original document and generate extractive data.

\paragraph{Abstractive Summarization.}
Abstractive summarization generates summaries using novel phrasing and sentence fusion or paraphrasing techniques \citep{shen-etal-2023-large, xiao-etal-2022-primera}. The seq2seq framework \citep{sutskever2014sequence} with encoder-decoder architectures based on RNNs \citep{chung2014empirical, hochreiter1997long} and Transformers \citep{vaswani2017attention} are dominant in this field. Recently, there has been a surge in prompting LLMs such as GPT \citep{brown2020language}. Studies like \citet{goyal2022news} have investigated the performance of GPT-3 and fine-tuned models, finding that the former is more preferred by humans despite having lower ROUGE scores. \citet{zhang2023summit} iteratively refines summaries through self-evaluation and feedback, exploring the use of knowledge and topic extractors to enhance summary faithfulness and controllability.
\citet{liu2023learning} finds that LLMs generate summaries preferred by humans and proposes improving PLMs using LLMs as references through supervised fine-tuning and contrastive learning.
In this paper, we also utilize the zero-shot summarization capability of LLMs to comprehensively understand entire documents and generate abstractive data. However, we find that LLM-generated summaries alone are not optimal, and incorporating more diverse data better aligns with human preferences.
% We then mix this abstractive data with extractive data to pretrain PLMs for generating domain-general summaries.

\paragraph{Domain Adaptation Summarization} Domain adaptation summarization has been widely studied in low-resource settings \cite{yu2021adaptsum, balde2024medvoc, fabbri2020improving}.
\citet{gururangan2020don} demonstrates that domain- and task-adaptive pretraining consistently improves performance, though their work focuses on eight classification tasks and relies solely on pretraining with unlabeled data. WikiTransfer \cite{fabbri2020improving} extends domain adaptation to summarization by fine-tuning pretrained models on pseudo-summaries generated from Wikipedia data using ROUGE matching, which is suboptimal for abstractive summarization.
In this paper, we extend the concept of domain to encompass more refined human preferences, incorporating both ROUGE-based extractive methods and LLM-based abstractive methods to construct supervised training data, which introduces greater diversity and consistently enhances summarization performance.

\section{Conclusion}
% We observe a contradiction between automatic and human evaluation of PLMs, largely due to the low quality of raw datasets. To address this, we introduce a novel data construction method named DP to alleviate the scarcity of train data. We also propose a novel training framework that incorporates DP, Gaussian Resample, and Two Stages Training to align PLMs with human evaluation. Comprehensive experiments and ablation studies demonstrate the effectiveness of our framework and the importance of its various components across both automatic and human evaluation.

We propose a novel human summarization preference alignment framework \alignsum ~including Data Pymarid, Gaussian Resampling, and Two-stage Hierarchical Fine-Tuning to align PLMs with human preference. Experiments demonstrate the effectiveness of our framework and narrow the gaps between automatic and human evaluation of PLMs.

\section*{Limitations}

\paragraph{Dataset Diversity.}
% Due to the scarcity of high-quality text annotation data, our experiments were conducted solely on the Element-Aware dataset proposed by \citet{wang2023element}. The Element-Aware annotations for CNN/DailyMail and BBC Xsum are based on news elements, embodying only one type of user annotation pattern. On the other hand, Element-Aware contains only 200 samples, resulting in both the training and test sets comprising merely 100 samples each, indicating a limited train and test size. 

High-quality preference data acquisition is challenging due to the need for specialized and uniform annotation protocols, along with significant labor and time costs. These preferences are typically implicit and often reflect differences in writing styles, which complicates the annotation process.

Due to the scarcity of preference data, our experiments are limited to {\it CNN/DailyMail} and {\it BBC XSum} datasets, as they are the only twos with rewritten versions that reflect human preferences. However, this does not imply that our method is restricted to these datasets. If more high-quality preference datasets become available in the future, we are eager to extend our method to a broader range of datasets.

\paragraph{Language Model Usage.}
The zero-shot summarization capabilities of LLMs have shown impressive results, making them a seemingly ideal choice for generating summaries that align with human preferences. However, human summarization preferences are inherently implicit, requiring the design of extremely sophisticated prompts to elicit the desired responses from LLMs. This process is challenging and often uncontrollable in real-world scenarios.
At the same time, in-context learning is also unrealistic because the text length of the summarization task is much longer than other natural language tasks, and the upper limit of length is unpredictable.

In contrast, directly fitting implicit preferences using PLMs is a more efficient approach. This method offers irreplaceable advantages in terms of cost and resource consumption, making it a more practical solution for the summarization preference alignment.

\section*{Ethics Statement}
We utilize publicly available datasets and weight parameters for model training and data generation, all of which are accompanied by bibliographic citations, ensuring no ethical issues are involved.

\section*{Acknowledgements}
I would like to express my gratitude to the anonymous reviewers for their meticulous and diligent review efforts. This work is supported by the National Science and Technology Major Project 2023ZD0120703 and the China NSFC Projects (U23B2057, 62106142, 62176153 and 62120106006) and Shanghai Municipal Science and Technology Major Project (2021SHZDZX0102).

\bibliography{anthology,custom}
\bibliographystyle{acl_natbib}

\newpage

\appendix

% \begin{table}[]
% \centering
% \resizebox{\textwidth}{!}{
%     \begin{tabular}{@{}c|c|c@{}}
%     \toprule
%      &
%       CNN/DailyMail &
%       BBC Xsum \\ \midrule
%     Data Feature &
%       \begin{tabular}[c]{@{}c@{}}Avg. summary length of words/sentences:\\ 51.08/2.71\end{tabular} &
%       \begin{tabular}[c]{@{}c@{}}Avg. summary length of words/sentences:\\ 23.33/1.00\end{tabular} \\ \midrule
%     User Prompt &
%       Summarize the article in three sentences around 60 words. &
%       Summarize the article in one sentence around 35 words. \\ \bottomrule
%     \end{tabular}%
% }
% \caption{\label{user-prompt} Data feature and user prompt for CNN/DailyMail and BBC Xsum.}
% \end{table}

\section{Detailed Theoretical Derivation: Why Hierarchical Fine-tuning?}
\label{sec:theory}

% \begin{assumption}
% For hierarchical data $\{X, Y, Z\}$, data at the lower level of DP enhances the model's ability of the upper-level tasks, but data at the upper level impairs the model's ability of the lower-level tasks, i.e.,
% \begin{equation}
%     \begin{aligned}
%         &H(Z|J(p_{\theta} = p_{x;\theta})) 
%         > H(Z|J(p_{\theta} = p_{x,y;\theta})) \\
%         &~~~~~~~~~~~~~~~~~~~~~~~~~~~~~~~~> H(Z|J(p_{\theta} = p_{x,y,z;\theta})), \\
%         &H(Y|J(p_{\theta} = p_{x;\theta})) 
%         > H(Y|J(p_{\theta} = p_{x,y;\theta})), \\
%         &H(Y|J(p_{\theta} = p_{x,y;\theta})) < H(Y|J(p_{\theta} = p_{x,y,z;\theta})) \\
%         &H(X|J(p_{\theta} = p_{x;\theta})) 
%         < H(X|J(p_{\theta} = p_{x,y;\theta})) \\
%         &~~~~~~~~~~~~~~~~~~~~~~~~~~~~~~~~~< H(X|J(p_{\theta} = p_{x,y,z;\theta})). \\
%     \end{aligned}
% \end{equation}
% \end{assumption}

According to the Assumption \ref{assumption2} in main text, we have
\begin{equation}
    \begin{aligned}
        &G_{hy} - G_{hi} = \\
        &\left| H(Y,Z|J(p_{x,y,z;\theta})) - H(Y,Z|J(p_{x;\theta})) \right| \\
        &- | H(Y|J(p_{x,y;\theta})) - H(Y|J(p_{x;\theta})) | \\
        &- | H(Z|J(p_{x,y,z;\theta})) - H(Z|J(p_{x,y;\theta})) |\\
        =& ~H(Y,Z|J(p_{x;\theta})) - H(Y,Z|J(p_{x,y,z;\theta})) \\
        &+ H(Y|J(p_{x,y;\theta})) - H(Y|J(p_{x;\theta})) \\
        &+ H(Z|J(p_{x,y,z;\theta})) - H(Z|J(p_{x,y;\theta})) \\
        =& \left[ H(Y,Z|J(p_{x;\theta})) - H(Y|J(p_{x;\theta})) \right] \\
        &+ [H(Y|J(p_{x;\theta})) - H(Z|J(p_{x,y;\theta}))] \\
        &- [H(Y,Z|J(p_{x,y,z;\theta})) - H(Z|J(p_{x,y,z;\theta}))] \\
        =& ~[H(Z|Y,J(p_{x;\theta})) - H(Z|J(p_{x,y;\theta}))]\\
        &+ [H(Y|J(p_{x,y;\theta})) - H(Y|Z,J(p_{x,y,z;\theta}))] \\
        =& ~H(Y|J(p_{x,y;\theta})) - H(Y|Z,J(p_{x,y,z;\theta})) \\
        <&~0\\
    \end{aligned}
\end{equation}

\section{Detailed Experimental Setup}
\subsection{Human Preference Features of Element-Aware Dataset}\label{element-aware-instruction}

The annotators are required to adhere to two types of preferences \citep{wang2023element} when writing.
\paragraph{Macro Preference.} 
All news summaries must focus on the four dimensions: {\bf Fluency, Coherence, Consistency, and Relevance}.
\paragraph{Micro Preference.} 
All news summaries should have four essential core elements — {\bf Entity, Date, Event, and Result} — following the ``Lasswell Communication Model'' \citep{lasswell1948structure}. These elements must be faithful to the source document.

These preferences reflect professional implicit writing styles that are embedded within the text and are difficult to capture explicitly.

\subsection{LLM Inference Setting}\label{llm-inference}
The GPT-3 results are adapted from SumCoT \cite{wang2023element} and reevaluated using the \texttt{evaluate} package\footnote{https://github.com/huggingface/evaluate}. GPT-3.5 results are obtained via the OpenAI API with a temperature of 1. The LLaMA series results are inferred from pre-trained weights with a temperature of 0.6.

\subsection{Dataset Examples}\label{data_case}
We present examples of ED, AD, and HD in {\it CNN/DailyMail} (Table \ref{tab:cnncase}) and {\it BBC XSum} (Table \ref{tab:xsumcase}) datasets.

\subsection{Main Experiment Packages}\label{appendix:packages}
Table \ref{package-url} shows the links of pre-trained model weights and evaluation metrics used in this paper.

\begin{table}[htbp]
\resizebox{\columnwidth}{!}{%
\begin{tabular}{@{}c|c@{}}
    \toprule
    Model  & URL                                                  \\ \midrule
    BART-Large   & \begin{tabular}[c]{@{}c@{}}\href{https://huggingface.co/facebook/bart-large-cnn}{https://huggingface.co/facebook/bart-large-cnn}  \\    \href{https://huggingface.co/facebook/bart-large-xsum}{https://huggingface.co/facebook/bart-large-xsum} \end{tabular}     \\ \midrule
    BART-base   & \begin{tabular}[c]{@{}c@{}}\href{https://huggingface.co/ainize/bart-base-cnn}{https://huggingface.co/ainize/bart-base-cnn}  \\    \href{https://huggingface.co/Vexemous/bart-base-finetuned-xsum}{https://huggingface.co/Vexemous/bart-base-finetuned-xsum} \end{tabular}     \\ \midrule
    T5-Large     & \begin{tabular}[c]{@{}c@{}}\href{https://huggingface.co/kssteven/T5-large-cnndm}{https://huggingface.co/kssteven/T5-large-cnndm}  \\    \href{https://huggingface.co/kssteven/T5-large-xsum}{https://huggingface.co/kssteven/T5-large-xsum} \end{tabular}            \\ \midrule
    PEGASUS & \begin{tabular}[c]{@{}c@{}} \href{https://huggingface.co/google/pegasus-cnn\_dailymail}{https://huggingface.co/google/pegasus-cnn\_dailymail}\\ \href{https://huggingface.co/google/pegasus-xsum}{https://huggingface.co/google/pegasus-xsum}\end{tabular} \\ \midrule
    LLaMA2 & \href{https://huggingface.co/meta-llama/Llama-2-7b-chat-hf}{https://huggingface.co/meta-llama/Llama-2-7b-chat-hf} \\ \midrule
    LLaMA3 & \href{https://huggingface.co/meta-llama/Meta-Llama-3-8B-Instruct}{https://huggingface.co/meta-llama/Meta-Llama-3-8B-Instruct} \\
    \midrule Rouge-1/2/L & \href{https://huggingface.co/docs/evaluate/index}{https://huggingface.co/docs/evaluate/index} \\
    \midrule BERTScore & \href{https://github.com/Tiiiger/bert\_score}{https://github.com/Tiiiger/bert\_score} \\
    \bottomrule
    \end{tabular}%
}
\caption{\label{package-url} Links of pre-trained model weights and evaluation metrics used in the paper.}
\end{table}

\subsection{Selection of LLMs for AD Generation}\label{appendix:llm_ad}
In this paper, we use L{\footnotesize LaMA}-2-7B for generating AD, as Table \ref{tab:llm_abation} shows minimal improvement when using different LLMs, making L{\footnotesize LaMA}-2-7B a more efficient choice.

\begin{table}[ht]
  \centering
    %   \vspace{-0.15in}
    \resizebox{1\columnwidth}{!}{
    \begin{tabular}{c|ccc}
    
    \toprule

   \textbf{ Model}  &\textbf{R{\footnotesize OUGE}-1} & \textbf{R{\footnotesize OUGE}-2} & \textbf{R{\footnotesize OUGE}-L} \\
    \midrule
    L{\footnotesize LaMA}2-7B & 48.39 & 23.33 & 34.47 \\ 
    L{\footnotesize LaMA}3-8B & 48.47 & 22.83 & 33.15 \\ 
    L{\footnotesize LaMA}3-70B & 48.92 & 23.58 & 34.32 \\ 
    \bottomrule
    \end{tabular}%
    }
% \vspace{-0.1in}
\caption{Ablation study of using different LLMs to generate AD in \alignsum. }
\label{tab:llm_abation}
\end{table}%

\section{Reference-free Human Evaluation}\label{appendix:reference-free-human}
We recruit the same annotators (one Ph.D. student and two Master's students) to evaluate 25 randomly selected \textit{BBC XSum} samples based on four criteria: coherence, consistency, fluency, and relevance (rated 1-5,~\citet{fabbri2021summeval}).
We select GPT-3, GPT-3 (\textit{w/SumCoT}) and the original summary as baselines, for each sample, we ask the annonators to score from four aspects, and average their rating as the final score.
Table \ref{tab:performance_comparison} presents the average score of the 25 randomly selected samples.

\begin{table}[ht]
\centering
\resizebox{1\columnwidth}{!}{
\begin{tabular}{lcccc}
\toprule
\textbf{Method} & \textbf{Coherence} & \textbf{Consistency} & \textbf{Fluency} & \textbf{Relevance} \\ \midrule
Original Summary & 3.4 & 3.2 & 4.2 & 3.16 \\ 
\alignsum & 4.0 & 3.72 & 4.48 & 3.88 \\ 
GPT-3 & 4.06 & 3.6 & 4.44 & 3.8 \\ 
GPT-3 (\textit{w/SumCoT}) & 4.32 & 4.4 & 4.52 & 3.96 \\ \bottomrule
\end{tabular}
}
\caption{Reference-free human evaluation results across different methods.}
\label{tab:performance_comparison}
\end{table}

GPT-3 (\textit{w/SumCoT}) outperforms \alignsum ~in a reference-free setting, but this does not contradict our conclusions, as our goal is to align with the specific human preferences reflected in the human-annotated data. On the other hand, \alignsum ~outperforms the original summary and performs comparably to GPT-3, demonstrating its ability to generate high-quality summaries while aligning with human preferences.

\section{Supplementary Experimental Results}
\subsection{BART(w/ full DP) Results on Random Samples}\label{random_exp}
We randomly split the Element-Aware Dataset five times, Table \ref{exp:cnn5} and Table \ref{exp:xsum5} show the automatic evaluation on \textit{CNN/DailyMail} and \textit{BBC XSum}, respectively.

\begin{table}[htbp]
\resizebox{\columnwidth}{!}{%
\begin{tabular}{c|cccc}
\hline
\textbf{Exp} & \textbf{R1} & \textbf{R2} & \textbf{RL} & \textbf{BERTScore} \\ \hline
random1      & 47.82       & 22.34       & 33.02       & 0.9050             \\
random2      & 51.82       & 29.44       & 38.04       & 0.9124             \\
random3      & 48.08       & 23.24       & 33.81       & 0.9045             \\
random4      & 48.28       & 22.98       & 32.88       & 0.9036             \\
random5      & 48.17       & 22.58       & 33.06       & 0.9033             \\ \hline
Mean         & 48.83      & 24.12      & 34.16      & 0.9058             \\ \hline
\end{tabular}%
}
\caption{Experiments on \textit{CNN/DailyMail}}\label{exp:cnn5}
\end{table}

\begin{table}[htbp]
\resizebox{\columnwidth}{!}{%
\begin{tabular}{c|cccc}
\hline
\textbf{Exp} & \textbf{R1} & \textbf{R2} & \textbf{RL} & \textbf{BERTScore} \\ \hline
random1      & 45.69       & 20.71       & 34.42       & 0.8992             \\
random2      & 43.25       & 18.64       & 32.22       & 0.8972             \\
random3      & 41.34       & 16.88       & 32.12       & 0.8959             \\
random4      & 41.25       & 16.26       & 29.58       & 0.8943             \\
random5      & 40.35       & 16.26       & 29.84       & 0.8944             \\ \hline
Mean         & 42.38      & 17.75       & 31.64      & 0.8962             \\ \hline
\end{tabular}%
}
\caption{Experiments on \textit{BBC XSum}}\label{exp:xsum5}
\end{table}

\subsection{Case Study}\label{case_study}
We show the "Win", "Equal" and "Fail" case on \textit{CNN/DailyMail} in Table \ref{tab:win-case}, Table \ref{tab:equal-case} and Table \ref{tab:fail-case} respectively.
% \usepackage{booktabs}
\begin{table*}[tbp]
\resizebox{\textwidth}{!}{
\begin{tabular}{p{4cm} | p{12cm}}
\toprule
\textbf{BART(w/ full DP)}                             & An elderly woman's body was found mummified in a house in San Francisco's Richmond area, believed to be owned by a hoarder. The body was discovered on Saturday, and police are investigating how it got to this point. \\ \midrule
\multicolumn{1}{l|}{\textbf{Golden Reference}} & In one of the worst hoarding cases in recent years, San Francisco police found the body of an elderly woman in a hoarder's home on Saturday, with trash debris piled up in the room.  Police are working to investigate the cause of the case. \\ \hline
\textbf{GPT-3 w/CoT}                           &   Police in San Francisco have discovered the mummified body of an elderly woman in a house said to be owned by a hoarder.   The property in the Richmond area is said to have been crawling with rats, black widow spiders, dog feces and 300 bottles of urine.   The remains, which were removed from the house on Saturday, are believed to be that of Anna Ragin who lived with her 65-year-old daughter Carolyn.  Neighbors told the San Francisco Chronicle the mother had not been seen for years.   According to records another man, Archibald Ragin, lived at the property but died 15 years ago at the age of 101.   Carolyn, said to be a retired Pacific Bell worker, was taken to hospital after her mother's body was removed from the house.  Supervisor Eric Mar, a representative in charge of the area, told the Chronicle: \"The police captains I've spoken with tell me this is the worst case of hoarding they have ever seen.  \" He was concerned for the health and safety of the residents but questioned how the situation in the property could have gotten to such a point.  When city workers arrived to clean up the body on Wednesday, they reportedly struggled to get in the door because the house was filled with so much debris.   The mess was said to have included mold and piles of trash that reached the ceiling.   They were forced to call in professional workers.  The Chronicle quotes a firefighter saying the body was lightweight and wrapped in a blanket.   It is not clear where it was discovered.  \\ \bottomrule
\end{tabular}
}
\caption{The ``Win'' Case. BART(w/ full DP) generates a summary of similar length compared to the golden reference, whereas GPT-3 w/CoT produces an excessively long summary.}\label{tab:win-case}
\end{table*}

\begin{table*}[htbp]
\resizebox{\textwidth}{!}{
\begin{tabular}{p{4cm} | p{12cm}}
\toprule
\textbf{BART(w/ full DP)}                             &  \textcolor{red}{Jason Rezaian}, a  \textcolor{red}{Washington Post bureau chief in Tehran}, has been \textcolor{green}{held in jail in Iran} for nearly nine months on unspecified charges of economic spying. His wife, Yeganeh Salehi, was released on bail in October, but he has been denied access to proper legal representation and has been subjected to long interrogations. \\ \midrule
\multicolumn{1}{l|}{\textbf{Golden Reference}} & Police \textcolor{green}{arrested} \textcolor{red}{Jason Rezaian}, the \textcolor{red}{Washington Post's Tehran bureau chief}, on \textcolor{blue}{July 22} on unknown charges, and \textcolor{yellow}{did not make the charges public until last week, more than four months after the arrest}. \\ \hline
\textbf{GPT-3 w/CoT}                           &   \textcolor{red}{Jason Rezaian}, the \textcolor{red}{Washington Post's bureau chief in Tehran}, \textcolor{green}{has been in jail} in Iran for nearly nine months on unspecified charges.  Last week, it was revealed that he will be tried soon on espionage charges.  The Washington Post has condemned the charges as \"absurd\" and called for Rezaian's release.  Rezaian has been denied bail and has been subjected to long interrogations.  His wife was released on bail in October.  \\ \bottomrule
\end{tabular}
}
\caption{The ``Equal'' case. We represent the entities, dates, events, and results in  \textcolor{red}{red}, \textcolor{blue}{blue}, \textcolor{green}{green} and \textcolor{yellow}{yellow}. It is clear that Info$_1=$ Info$_2 =3$.}\label{tab:equal-case}
\end{table*}

\begin{table*}[tbp]
\resizebox{\textwidth}{!}{
\begin{tabular}{p{4cm} | p{12cm}}
\toprule
\textbf{BART(w/ full DP)}                             & On \textcolor{blue}{Wednesday}, \textcolor{red}{Andrea Atzeni} will ride \textcolor{red}{terror} in the 1,000 Guineas at Newmarket, who finished fourth in the group one Cheveley park stakes over six furlongs at the Rowley mile last September. \textcolor{red}{David Simcock}, trainer of terror, believes she has been solid in training and is not a bad filly, but it’s difficult to know where she stands fitness-wise until she gets into a race scenario. \\ \midrule
\multicolumn{1}{l|}{\textbf{Golden Reference}} & Jockey \textcolor{red}{Andrea Atzeni} will ride \textcolor{red}{Terror} in \textcolor{blue}{Wednesday}'s \textcolor{red}{Landwades Nell Gwyn Stakes at Newmarket}. Terror, trained by \textcolor{red}{David Simcock}, is one of the \textcolor{red}{contenders for the 1,000 Guineas}. \textcolor{green}{This race will be a great opportunity for Atzeni to publicize the horse's chances in the Classic}. \\ \hline
\textbf{GPT-3 w/CoT}                           &   Jockey \textcolor{red}{Andrea Atzeni} will ride the horse \textcolor{red}{Terror} in the \textcolor{red}{Landwades Nell Gwyn Stakes at Newmarket} on \textcolor{blue}{Wednesday}.  Terror is trained by \textcolor{red}{David Simcock} and is one of the \textcolor{red}{contenders for the 1,000 Guineas}.  \textcolor{green}{This race will be a good opportunity for Atzeni to promote the horse's chances in the classic}.  \\ \bottomrule
\end{tabular}
}
\caption{The ``Fail'' case. We represent the entities, dates, events, and results in  \textcolor{red}{red}, \textcolor{blue}{blue}, \textcolor{green}{green} and \textcolor{yellow}{yellow}. It is clear that Info$_1 = 4 <$ Info$_2 = 7$.}\label{tab:fail-case}
\end{table*}

\begin{table*}[tbp]
\resizebox{\textwidth}{!}{
\begin{tabular}{p{4cm} | p{12cm}}
\toprule
\textbf{Document}                             & conrad clitheroe and gary cooper , both from stockport , and expat neil munro were reportedly taking notes near fujairah airport , 80 miles from dubai , when they were arrested in february . relatives were told they were held for `` national security '' reasons . the men insisted they did not take photographs . the abu dhabi hearing is due on monday . mr clitheroe , 54 , and mr cooper 45 , were visiting their friend mr munro , who was born in manchester , when they were arrested on 22 february by an off-duty police officer who had seen them monitoring planes from a car . they were near fujairah airport , where older and rarer aircraft can be seen . a local police official said the men had been taking photographs near an airport and were using a telescope . the men are expected to argue their actions were misinterpreted and are understood to be hoping to be granted bail . \\ \midrule

\multicolumn{1}{l|}{\textbf{ED}} & mr clitheroe , 54 , and mr cooper 45 , were visiting their friend mr munro , who was born in manchester , when they were arrested on 22 february by an off-duty police officer who had seen them monitoring planes from a car . \\ \hline

\multicolumn{1}{l|}{\textbf{AD}} & Three British men, Conrad Clitheroe and Gary Cooper from Stockport and Neil Munro from Manchester, were arrested near Fujairah Airport in February for taking notes and using a telescope, with their lawyer expected to argue that their actions were misinterpreted and they are hoping to be granted bail. \\ \hline

\textbf{HD}                           &   Three men were arrested for taking notes and taking photographs near fujairah airport in February, they hope to be granted bail for being misinterpreted.  \\ \bottomrule

\end{tabular}
}
\caption{Case of ED, AD and HD in \textit{BBC XSum}}
\label{tab:xsumcase}
\end{table*}

\begin{table*}[tbp]
\resizebox{\textwidth}{!}{
\begin{tabular}{p{4cm} | p{12cm}}
\toprule
\textbf{Article}                             & A married software executive who drugged a female employee in order to take naked pictures of her on a business trip has been jailed. Sexual predator Henri Morris was told he would serve 10 years behind bars for his 'calculated and choreographed' crime.  The 67-year-old was caught in an FBI sting after investigators were approached by one of his victims in 2012. Henri Morris, 67, was jailed for 10 years after admitting drugging a female employee during a business trip in order to take naked photos of her She told them that her drink was spiked by the married businessman after they traveled together from Houston, Texas, to New Jersey for work. The woman said when she woke up she was naked and her boss was standing over her and taking pictures on his mobile phone. The FBI arrested Morris at Bush County Airport after the woman, who has not been named, covertly worked with them. When his bags were search they found his 'kit,' which included strong sedatives and Viagra. A task force was set up to probe Morris and in total they found eight female employees who claimed to have been abused by him. His defense team initially tried to argue that the women all willingly drank to excess with clients and that no abuse had ever occurred. Ahead of his trial Morris pleaded guilty to a single charge of drugging and abusing a female employee, in exchange for prosecutors dropping other abuse charges. MarriedMorris of Houston, Texas, was the president of Edible Software Solutions  'There is no way to adequately express my remorse and my abject humiliation,' Morris said during a 10 minute address at his sentencing hearing, according to the Houston Chronicle. 'I apologize to anyone who I have hurt and beg for their forgiveness.'   US District Judge Melinda Harmon sentenced him to 10 years in federal prison without the possibility of parole. She also reportedly ordered that he spend the rest of his life under the supervision of the US Bureau of Prisons, according to the newspaper. \\ \midrule

\multicolumn{1}{l|}{\textbf{ED}} & Henri Morris, 67, was jailed for 10 years after admitting drugging a female employee during a business trip in order to take naked photos of her She told them that her drink was spiked by the married businessman after they traveled together from Houston, Texas, to New Jersey for work. \\ \hline

\multicolumn{1}{l|}{\textbf{AD}} & A married software executive, Henri Morris, has been sentenced to 10 years in federal prison for drugging and sexually assaulting a female employee during a business trip. Morris spiked the woman's drink with sedatives and took naked pictures of her while she was unconscious. The FBI arrested Morris after one of his victims came forward and the task force found eight other female employees who had been abused by him. \\ \hline

\textbf{HD}                           &   Henri Morris, 67, was a married software executive, he was jailed for 10 years for drugging a female employee during a business trip in order to take naked photos of her, without the possibility of parole.  \\ \bottomrule

\end{tabular}
}
\caption{Case of ED, AD and HD in \textit{CNN/DailyMail}}
\label{tab:cnncase}
\end{table*}

\end{document}